\documentclass[conference]{IEEEtran}
\IEEEoverridecommandlockouts
\usepackage{cite}
\usepackage{amsmath,amssymb,amsfonts,bm}
\usepackage{algorithm}
\usepackage[noend]{algpseudocode}
\usepackage{booktabs,array,tabularx}
\usepackage{xcolor}
\usepackage{tikz}
\usepackage{graphicx} 
\usetikzlibrary{positioning,arrows.meta}
\usepackage{amsthm}

\def\BibTeX{{\rm B\kern-.05em{\sc i\kern-.025em b}\kern-.08em
T\kern-.1667em\lower.7ex\hbox{E}\kern-.125emX}}

\newcommand{\tht}{\boldsymbol{\theta}}
\newcommand{\ttht}{\tilde{\boldsymbol{\theta}}}
\newcommand{\btht}{\bar{\boldsymbol{\theta}}}
\newcommand{\n}{\mathbf{n}}
\newcommand{\m}{\mathbf{m}}
\newcommand{\vv}{\mathbf{v}}
\newcommand{\g}{\mathbf{g}}
\newcommand{\HH}{\mathbf{H}}
\newcommand{\I}{\mathbf{I}}
\newcommand{\E}{\mathbb{E}}

\newcommand{\Gauss}{\mathcal{N}}
\newcommand{\Hset}{\mathcal{H}}
\newcommand{\Bset}{\mathcal{B}}
\newcommand{\Cset}{\mathcal{C}}
\newcommand{\Pset}{\mathcal{P}}
\newcommand{\Qset}{\mathcal{Q}}
\newcommand{\D}{\mathcal{D}}

\algrenewcommand\algorithmiccomment[1]{\hfill{\footnotesize\(\triangleright\) #1}}

\definecolor{hon}{HTML}{2A78D6}
\definecolor{byz}{HTML}{EB6834}
\definecolor{res}{HTML}{1BAF7A}
\definecolor{alt}{HTML}{4A3AA7}
\definecolor{yel}{HTML}{EDA100}

\title{FedSharp: Byzantine Robust Decentralized Federated Learning Under Noisy Channel}
\author{\IEEEauthorblockN{Ojaswi Sinha\IEEEauthorrefmark{1}, Amin Farajzadeh\IEEEauthorrefmark{2}, \textit{Member}, \textit{IEEE}, Animesh Yadav\IEEEauthorrefmark{1}, \textit{Senior Member}, \textit{IEEE}\\ and Melike Erol-Kantarci\IEEEauthorrefmark{2}, \textit{Fellow}, \textit{IEEE}}
\IEEEauthorblockA{\IEEEauthorrefmark{1}School of EECS, Ohio University, Athens, OH, USA, \IEEEauthorrefmark{2}School of EECS, University of Ottawa, Ottawa, ON, Canada}
Email: \IEEEauthorrefmark{1}\{os675524, yadava\}@ohio.edu, \IEEEauthorrefmark{2}\{amin.farajzadeh, melike.erolkantarci\}@uottawa.ca
}

\begin{document}
\maketitle
\begin{abstract}
Decentralized federated learning (DFL) eliminates the central aggregation server but remains vulnerable to data heterogeneity, Byzantine attacks, and channel noise. We show that noise accumulates as models propagate across nodes, becoming increasingly significant as the learning rate decays. Moreover, accumulated noise enables Byzantine attackers to exploit loss-based model selection by mimicking honest model averages. To address these challenges, we propose FedSharp (screened, hop-aware, averaged, ranked, proximal), a robust DFL framework. In this approach, each node maintains models from its $2b$ predecessors, ensuring at least $b$ honest candidates when at most $b$ nodes are Byzantine. FedSharp screens suspicious models using coordinate-wise median deviations, averages the $b$ lowest-loss candidates to reduce fresh channel-noise variance by a factor of $b$, and applies noise-adaptive gradient-norm regularization and proximal training to mitigate accumulated noise and non-IID drift. Experiments on a 100-node ring with 33 Byzantine nodes show that FedSharp achieves 93.5–94.9$\%$ accuracy on MNIST, compared with 29.4–38.1$\%$ for a state-of-the-art single-model-selection benchmark, and 32.8–48.8$\%$ on CIFAR-10, compared with 10.0–30.5$\%$, across Gaussian and hidden attacks under IID and non-IID data. On MNIST, FedSharp also achieves accuracy within 1.3 percentage points of, and under the hidden attack above, the benchmark operating without channel noise.
\end{abstract}


\section{Introduction}
Federated learning (FL) facilitates collaborative model training without the necessity of sharing raw local data, with FedAvg serving as the foundational framework characterized by local optimization followed by centralized aggregation \cite{mcmahan_fedavg,mel_fl_icc24}. In contrast, decentralized federated learning (DFL) removes the need for a central coordinator by enabling direct model exchanges through peer-to-peer, gossip, or ring-based communication \cite{lian2017, balancingcomm,fedisp,pfedgame}. This decentralized approach mitigates the risk of a single point of failure and is well suited for Internet-of-Things (IoT) and device-to-device (D2D) networks~\cite{xing2021}. However, it introduces training challenges, particularly non-independent and identically distributed (non-IID) client data, which can cause local model drift and degrade overall learning performance~\cite{hsu2019,zhao_non_iid,fedprox,scaffold}.

In addition to data heterogeneity, DFL is susceptible to the following two sources of model degradation: i) Byzantine attacks and ii) communication channel noise. Byzantine clients may, intentionally, transmit corrupted model states, necessitating the development of robust aggregation and Byzantine-tolerant methodologies \cite{Byzantine_Tolerant_GD,geometric_median,fall_of_empires,analyzing_fl_adversarial,basil}. Conversely, unreliable wireless communication links, unintentionally, can randomly distort otherwise valid model parameters, prompting the need for channel noise-robust strategies \cite{unreliablecomm,wireless_edge_fl,robustwireless, NoiseComm}. Although Byzantine attacks and channel noise arise from fundamentally different mechanisms, their effects can interact and compound as models propagate across decentralized networks, particularly under non-IID data distributions. Understanding these interactions and developing effective mitigation strategies are therefore essential for robust DFL.

Motivated by this challenge, this paper jointly investigates the impact of Byzantine attacks, channel noise, and non-IID data and proposes FedSharp (screened, hop-aware, averaged, ranked, proximal), a novel framework designed to mitigate their compounded effects. To the best of our knowledge, this is the first study to systematically investigate the interplay of all three impairments and address them within a unified DFL framework. The main contributions are summarized as follows:
\begin{itemize}
    \item We propose FedSharp, which integrates Byzantine-resilient model screening, loss-based aggregation, hop-aware noise regularization, and proximal optimization to jointly mitigate Byzantine attacks, accumulated channel noise, and non-IID model drift.
    \item We evaluate FedSharp on MNIST and CIFAR-10 under IID and non-IID data, demonstrating its robustness under combined Byzantine attacks and channel noise.
\end{itemize}
\section{Related Work}
Data heterogeneity remains a fundamental challenge in federated learning (FL), as non-IID data can cause local model drift, slowing convergence and degrading learning performance \cite{zhao_non_iid,amin_non_iid,mel_non_iid}. In centralized FL, FedProx \cite{fedprox} employs proximal regularization to limit local model deviations, while SCAFFOLD \cite{scaffold} uses control variates to mitigate client drift. However, their reliance on centralized coordination limits their direct applicability to decentralized federated learning (DFL). In DFL, clients exchange models directly over peer-to-peer networks with different communication topologies \cite{beltran_dfl_survey,balancingcomm,fedisp,pfedgame}. Consequently, learning performance depends on both local data distributions and model propagation across the network \cite{chung_dfl_non_iid}. Existing approaches, such as PENS \cite{pens} and knowledge-distillation-based methods \cite{kd_fl_guide}, leverage performance-based peer selection and knowledge transfer, respectively. Nevertheless, preserving knowledge across heterogeneous clients without global coordination remains challenging \cite{beltran_dfl_survey,chung_dfl_non_iid}.

Byzantine attacks introduce another challenge, as malicious clients can transmit arbitrary updates that disrupt collaborative learning \cite{Byzantine_Tolerant_GD}. Robust aggregation methods, including geometric-median-based approaches \cite{geometric_median}, mitigate corrupted updates but remain vulnerable to sophisticated attacks \cite{fall_of_empires,analyzing_fl_adversarial}. Non-IID data further complicates Byzantine detection because legitimate heterogeneous updates may resemble malicious ones. To address this, GAS \cite{gas} employs gradient splitting to improve robustness under data heterogeneity. In DFL, Basil \cite{basil} introduces memory-assisted, performance-based model selection in a sequential ring, while BALANCE \cite{balance} filters suspicious neighboring models using local similarity. Bristle \cite{bristle} combines transfer learning and performance-based model integration to address Byzantine attacks under non-IID data.

Communication noise further degrades DFL performance by perturbing exchanged model parameters. Earlier studies investigated learning over noisy wireless channels and developed mechanisms to improve robustness against communication impairments \cite{wireless_edge_fl,robustwireless}. In particular, \cite{NoiseComm} proposed expectation-based and worst-case optimization approaches to mitigate stochastic transmission errors. More recently, \cite{li_imperfect_dfl} investigated error accumulation during decentralized aggregation and proposed adaptive aggregation based on network topology and channel conditions. Similarly, DFL with global momentum (DFLGM) \cite{li_dflgm} combines differential model exchange and global momentum to mitigate accumulated channel noise under heterogeneous data distributions. Collectively, these studies highlight the importance of robust model selection, heterogeneity-aware optimization, and communication-noise mitigation for reliable DFL.

\begin{figure}[ht]
    \centering
    \includegraphics[width=\linewidth]{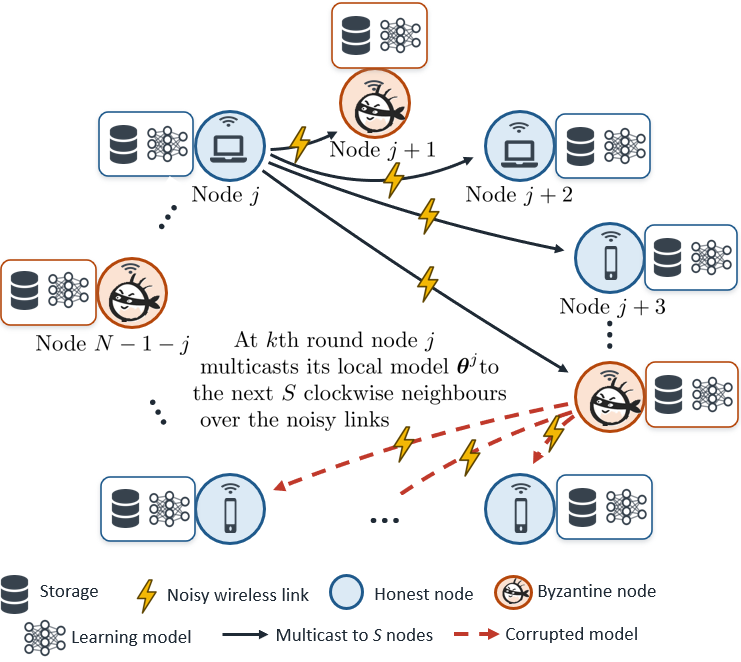}
    \caption{The DFL framework of $N$ nodes arranged on a ring topology.}
    \label{fig:system_model}
\end{figure}
\section{System Model}
\subsection{Network Topology and Decentralized Operation}
We consider a DFL system consisting of a set $\mathcal{N}=\{1,2,\ldots,N\}$ of $N$ edge clients or nodes learning collaboratively a global model $\boldsymbol{\theta}\in \mathbb{R}^d$, where $d$ is the size of the model vector, in the absence of a centralized server. The nodes are arranged in a directed ring topology as illustrated in Fig.~\ref{fig:system_model} and are indexed clockwise as $\{1,2,\ldots,N\}$, and the communication among them follows the clockwise direction. A set $\Bset$ of at most $b$ nodes is Byzantine, and the remaining set $\Hset$ is honest. The Byzantine nodes are placed uniformly at random on the ring. Honest node $i$ holds a local dataset $\D_i$ and the loss $\ell_i(\tht;\zeta)$ of model $\tht\in\mathbb R^d$ on a mini-batch $\zeta\subset\D_i$. 

\subsection{Model Training}
Training proceeds in rounds indexed by $k=1,2,\dots$. In each round $k$, every node is activated once, in ring order. When node $j$ is active it updates its model $\tht^{(j)}$ and multicasts it to its $S$ successors as depicted in Fig.~\ref{fig:system_model}. Consequently, each node stores a candidate set $\Cset_i$ with the latest models received from each of its $S$ predecessors. The overall goal is to minimize the following average loss of the honest nodes in the presence of Byzantine and channel noise:
\begin{equation}
\min_{\tht\in\mathbb R^d}\ F(\tht)=\frac{1}{|\Hset|}\sum_{i\in\Hset}f_i(\tht),\text{ where } f_i(\tht)=\E_{\zeta\subset\D_i}[\ell_i(\tht;\zeta)], \label{eq:obj}
\end{equation}
where $|\cdot|$ and $\mathbb{E}[\cdot]$ represent cardinality of the set and expectation operator, respectively. 

\subsection{Channel Model}
We assume that models are transmitted over noisy communication links. When node $j$ broadcasts $\tht^{(j)}$, receiver node $i$ obtains its noisy copy as
\begin{equation}
\ttht_{i,j}=\tht^{(j)}+\n_{i,j},\qquad \n_{i,j}\sim\Gauss(\mathbf 0,\sigma^2\I_d), \label{eq:channel}
\end{equation}
where $\n_{i,j}$ is the noise on the link from node $j$ to node $i$ and is iid across links, receivers, and transmissions. The variance $\sigma^2$ captures the residual error after equalization of an analog transmission~\cite{zhu2020,amiri2020}. Senders are authenticated, so each candidate set contains at most one received model vector per sender identity.
 
\subsection{Threat Model}
Byzantine nodes are omniscient as they know the honest models and may collude. Each may send one of the following vectors: (i) \emph{Gaussian attack}, where honest model vector is replaced by a random vector and (ii) \emph{hidden attack}, where model vector is replaced by the honest mean plus a large shift on a single parameter~\cite{mhamdi2018}. In this threat model, Byzantine nodes cannot forge identities or create fake ones. Hidden attackers behave honestly during the first 20 rounds.
 
\section{Problem Statement}\label{sec:problem}
An activated honest node $i$ combines the noisy models in its candidate set $\Cset_i$ into a starting model and then trains it on its local data $\D_i$. Two impairments affect this process and reinforce each other. First, since every model is built from noisy received versions of earlier models, link noise accumulates over hops: near a minimizer, a model retains the noise of roughly its last $1/(2\eta h)$ hops, where $\eta$ is the learning rate and $h$ the local curvature, so that with $q$ averaged received model vectors its noise variance settles at $v^\ast\approx\sigma^2/(2q\eta h)$. This noise floor rises as $\eta$ decays, and averaging $q$ received model vectors lowers it $q$-fold. Second, Byzantine model vectors bias the model in proportion to the weight $f$ that the aggregation gives them. Averaging more received model vectors thus reduces the noise but may admit more attackers, and noise makes this worse: a hidden attacker that sends the mean of the honest models carries less accumulated noise than any honest model, so a loss-based rule tends to prefer it. Since a candidate set of $S$ senders contains at least $S-b$ honest model vectors, at most $q=S-b$ received model vectors can be averaged safely. In particular, at each active honest node $i$, we seek an aggregation rule $\btht_i=\mathcal A(\Cset_i;\D_i)$ and a local update rule $\tht^{(i)}=\mathcal U(\btht_i;\D_i)$ that meet four design objectives:
\begin{enumerate}
\item[(O1)] honest model vectors are guaranteed to be available in every candidate set, for any placement of at most $b$ attackers;
\item[(O2)] the attacker weight $f$ is kept small, in particular for the hidden attack, which can win a loss-based ranking;
\item[(O3)] the fresh noise per hop, and hence the accumulated noise $v^\ast$, is reduced by averaging $q\gg1$ received model vectors;
\item[(O4)] the local update is robust to the residual noise and limits drift toward the node's own classes;
\end{enumerate}
so as to minimize the excess of the training objective \eqref{eq:obj} caused by noise, attacks and drift.
 
\section{FedSharp-Based Robust Decentralized Learning}\label{sec:FedSharp}
 
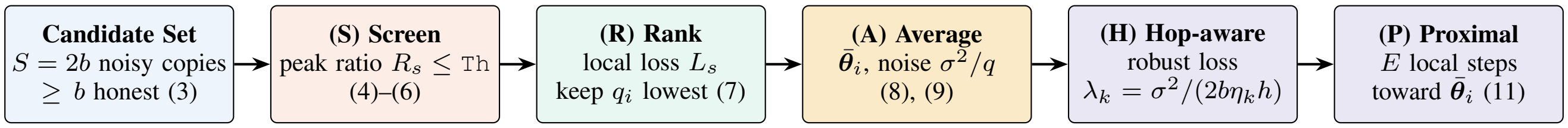
\begin{figure*}[t]\centering
\begin{tikzpicture}[box/.style={draw, rounded corners=2pt, align=center, font=\footnotesize, minimum height=13mm, text width=24mm, inner sep=2pt},
  arr/.style={-{Stealth[length=2mm]}, thick}]
\node[box, fill=hon!8] (w) {\textbf{Candidate Set}\\$S=2b$ noisy copies\\$\ge b$ honest \eqref{eq:q}};
\node[box, fill=byz!12, right=4mm of w] (p) {\textbf{(S) Screen}\\peak ratio $R_s\le \texttt{Th}$\\\eqref{eq:median}--\eqref{eq:pass}};
\node[box, fill=res!10, right=4mm of p] (r) {\textbf{(R) Rank}\\local loss $L_s$\\keep $q_i$ lowest \eqref{eq:keep}};
\node[box, fill=yel!22, right=4mm of r] (a) {\textbf{(A) Average}\\$\btht_i$, noise $\sigma^2/q$\\\eqref{eq:avg}, \eqref{eq:cov}};
\node[box, fill=alt!12, right=4mm of a] (h) {\textbf{(H) Hop-aware}\\robust loss\\$\lambda_k=\sigma^2/(2b\eta_kh)$};
\node[box, fill=alt!12, right=4mm of h] (t) {\textbf{(P) Proximal}\\$E$ local steps\\toward $\btht_i$ \eqref{eq:local}};
\draw[arr] (w)--(p); \draw[arr] (p)--(r); \draw[arr] (r)--(a); \draw[arr] (a)--(h); \draw[arr] (h)--(t);
\end{tikzpicture}
\caption{FedSharp at an active honest node. Screening and ranking remove attackers that look different; averaging reduces channel noise; the hop-aware robust loss and the proximal term shape the local training.}
\label{fig:pipeline}
\end{figure*}

Now, we describe our proposed approach FedSharp, which counters both Byzantine attacks and noisy channel links. The approach is applicable in both IID and non-IID data settings. The algorithm is presented in Algorithm~\ref{alg:sharp}, which consists of five major steps as shown in Fig.~\ref{fig:pipeline}.
 
\subsection{Candidate Set and Honest-Copy Guarantee}
This part achieves O1. Each node stores the latest copy \eqref{eq:channel} from each of its $S=2b$ predecessors. For any placement of at most $b$ Byzantine nodes, a full candidate set (reached once each of the $S$ predecessors has transmitted, i.e., within the first round) of $S$ distinct sending nodes contains at least $|\Cset_i\cap\Hset|\ \ge\ S-b=b$ honest copies. The number of copies averaged is set to the number guaranteed to be honest,
\begin{equation} 
q_i=\min\{b,\ \max\{1,|\Cset_i|-b\}\}, \label{eq:q}
\end{equation}
so $q_i=b$ from the second round on (and already in the first round for nodes $S+1,\dots,N$). With $S=2b$ the attackers can, in the worst placement, form half of a candidate set. $S=2b+1$ with $q=b+1$ yields a strict honest majority.

Next, we describe the five steps of the proposed  FedSharp approach. 
\paragraph{Step-1: Peak Screening}\label{sec:screen}
This step is incorporated to handle hidden attacks and to achieve O2. In a hidden attack, the model vector changes one parameter by a large amount and hardly changes the loss, so it often ranks \emph{among the best}. FedSharp therefore first screens the shape of each sender's model vector deviation from the coordinate-wise median of the candidate set as
\begin{align}
[\m_i]_p&=\operatorname{median}_{s\in\Cset_i}[\ttht_{i,s}]_p,\qquad \vv_s=\ttht_{i,s}-\m_i, \label{eq:median}\\
R_s&=\frac{\sqrt d\,\|\vv_s\|_\infty}{\|\vv_s\|_2}\quad(R_s=0\ \text{if}\ \vv_s=\mathbf 0), \label{eq:peak}\\
\Pset_i&=\{s\in\Cset_i:\ R_s\le \texttt{Th}\},\qquad \texttt{Th}=c\sqrt{2\ln d}, \label{eq:pass}
\end{align}
where $[\mathbf{m}_i]_p$ and $[\ttht_{i,s}]_p$ denote the $p$th element of vectors $\mathbf{m}_i\in \mathbb{R}^d$ and $\ttht_{i,s}\in \mathbb{R}^d$. The median model vector $\m_i\in\mathbb R^d$ takes, for each parameter $p$, the median of the $|\Cset_i|$ received model vectors. Since at least half of the copies are honest, $\m_i$ stays close to the honest models, so $\vv_s\in\mathbb R^d$ measures how the received model vector from sender $s$ deviates from the honest nodes. The peak ratio $R_s\in[1,\sqrt d\,]$ compares the largest entry of $\vv_s$ with its root-mean-square entry; $c$ is a safety factor, $\texttt{Th}$ the threshold, and $\Pset_i\subseteq\Cset_i$ is the set of model vectors that pass the peak test. For a deviation spread over all $d$ parameters, such as Gaussian noise, $R_s\approx\sqrt{2\ln d}$; a single-parameter deviation gives $R_s$ up to $\sqrt d$. With $c=2.5$ and $\vv_s$ pure Gaussian, the false-alarm probability is at most $2d\,Q(\texttt{Th})\le 2d^{1-c^2}$, negligible for practical $d$, where $Q(\cdot)$ is the Gaussian tail function. The screen is applied to full candidate set (i.e., $|\Cset_i| = S$) only.
 
\paragraph{Step-2: Loss Ranking}
This step handles the Gaussian attack and achieves O2, especially for the models that pass the peak screen but have a high loss. Node $i$ draws a mini-batch $\zeta\subset\D_i$, computes the loss $L_s=\ell_i(\ttht_{i,s};\zeta)$ of each model vector $\ttht_{i,s}\, \forall s\in\Pset_i$, and keeps the $q=\min\{q_i,|\Pset_i|\}$ copies with the smallest loss as
\begin{equation}
\Qset_i=\{s\in\Pset_i:\ r_i(s)\le q\}, \label{eq:keep}
\end{equation}
where $r_i(s)=1+\big|\{s'\in\Pset_i:\ L_{s'}<L_s\}\big|$. Note that the Basil approach in~\cite{basil} is the special case of \eqref{eq:keep} for $q=1$.

\paragraph{Step 3: Averaging}
This step reduces the channel noise and achieves O3. The node $i$ averages the model vectors selected from the previous step as
\begin{equation}
\btht_i=\frac{1}{|\Qset_i|}\sum_{s\in\Qset_i}\ttht_{i,s}. \label{eq:avg}
\end{equation}
Each selected model vector carries its own link noise $\n_{i,s}\sim\Gauss(\mathbf0,\sigma^2\I_d)$, independent across copies. Averaging $q$ independent noises divides their variance by $q$, so, if the selection does not depend on the noise, the fresh noise in $\btht_i$ has covariance as
\begin{equation}
\operatorname{Cov}\Bigg(\frac{1}{q}\sum_{s\in\Qset_i}\n_{i,s}\Bigg)=\frac{\sigma^2}{q}\I_d, \label{eq:cov}
\end{equation}
where $\operatorname{Cov}(\cdot)$ is covariance operator. Because the ranking in Step~2 uses the received model vectors, \eqref{eq:cov} is an idealized benchmark. Since the selected model vectors are averaged with equal weights, the attacker weight in $\btht_i$ is $f_i=|\Qset_i\cap\Bset|/|\Qset_i|$, the fraction of Byzantine senders among them; Steps~1 and~2 keep it small. Together with the noise accumulation described in Section~\ref{sec:problem}, it shows that averaging $q=b$ received model vectors lowers the accumulated noise by a factor of up to $b$ compared with using a single received model vector.

\paragraph{Step-4: Hop-Aware Robust Loss}
This step makes local training robust to the noise that remains and achieves O4. The robust objective of~\cite{NoiseComm} adds a gradient-norm penalty weighted by the noise variance, which approximates the loss increase caused by a perturbation of the parameters. In the ring, however, a model carries the accumulated noise rather than the noise of one link, so FedSharp sizes the weight to it:
\begin{equation}
\lambda_k=\frac{\sigma^2}{2b\,\eta_k\,h}, \qquad \ell^{\mathrm r}_{i,k}(\tht)=\ell_i(\tht)+\lambda_k\|\nabla\ell_i(\tht)\|_2^2 , \label{eq:rla}
\end{equation}
whose gradient is $\g+2\lambda_k\HH\g$, with $\g=\nabla\ell_i$ and $\HH=\nabla^2\ell_i$. The Hessian--vector product $\HH\g$ is obtained by double backpropagation~\cite{drucker1992} without forming $\HH$. The curvature scale $h$ is a design constant, and $\lambda_k$ grows as $\eta_k$ decays, in step with $v^\ast$.

\paragraph{Step-5: Proximal Local Training}
This step limits drift on non-IID data. Starting from $\tht^{(0)}=\btht_i$, the node takes $E$ gradient steps on $\ell^{\mathrm r}_{i,k}(\tht)+\frac\mu2\|\tht-\btht_i\|^2$~\cite{fedprox}:
\begin{equation}
\tht^{(e+1)}=\tht^{(e)}-\eta_k\big(\g_e+2\lambda_k\HH_e\g_e+\mu(\tht^{(e)}-\btht_i)\big), \label{eq:local}
\end{equation}
where the first step reuses $\zeta$ and each later step draws a fresh mini-batch. The proximal term is zero at the first step and pulls the model back toward the average of $q$ neighbours, which jointly cover more classes than the node itself. The result $\tht^{(i)}=\tht^{(E)}$ is multicast to the next $S$ nodes. Algorithm~\ref{alg:sharp} summarizes one activation.

\subsection{Resulting Aggregation and Update Rules}
Steps 1--3 define the FedSharp aggregation rule and Steps 4--5 its local update rule, i.e., the two maps sought in Section~\ref{sec:problem}:
\begin{align}
\mathcal A(\Cset_i;\D_i)&=\frac{1}{|\Qset_i|}\sum_{s\in\Qset_i}\ttht_{i,s},\qquad \Qset_i\subseteq\Pset_i\subseteq\Cset_i, \label{eq:Arule}\\
\mathcal U(\btht_i;\D_i)&=\tht^{(E)},\quad \tht^{(0)}=\btht_i,\ \ \tht^{(e+1)}\ \text{given by \eqref{eq:local}}, \label{eq:Urule}
\end{align}
where $\Pset_i$ is the set that passes the peak screen \eqref{eq:pass}, $\Qset_i$ the $q$ lowest-loss model vectors in it \eqref{eq:keep}, and the update minimizes the hop-aware robust loss \eqref{eq:rla} with the proximal term. Algorithm~\ref{alg:sharp} summarizes one activation.

\begin{algorithm}[t]
\caption{FedSharp at honest node $i$ in round $k$}
\label{alg:sharp}
\begin{algorithmic}[1]
\Require Candidate set $\Cset_i$; data $\D_i$; $b,\sigma,h,\mu,E,\eta_k,\texttt{Th}$
\State $q_i\gets\min\{b,\max\{1,|\Cset_i|-b\}\}$ \Comment{\eqref{eq:q}}
\State $\Pset_i\gets\Cset_i$
\If{$|\Cset_i|=S$}
 \State compute $\m_i$ and $R_s$, $s\in\Cset_i$ \Comment{\eqref{eq:median}, \eqref{eq:peak}}
 \State $\Pset_i\gets\{s:R_s\le \texttt{Th}\}$ if nonempty \Comment{\eqref{eq:pass}}
\EndIf
\State draw $\zeta$; $L_s\gets\ell_i(\ttht_{i,s};\zeta)$, $s\in\Pset_i$
\State $\Qset_i\gets$ the $\min(q_i,|\Pset_i|)$ lowest $L_s$ \Comment{\eqref{eq:keep}}
\State $\btht_i\gets$ average of $\Qset_i$; $\tht\gets\btht_i$ \Comment{\eqref{eq:avg}}
\State $\lambda_k\gets\sigma^2/(2b\,\eta_k h)$ \Comment{\eqref{eq:rla}}
\For{$e=0,\dots,E-1$}
 \State $\zeta_e\gets\zeta$ if $e=0$, else draw $\zeta_e$; update $\tht$ by \eqref{eq:local}
\EndFor
\State multicast $\tht^{(i)}\gets\tht$ to the next $S$ nodes \Comment{\eqref{eq:channel}}
\end{algorithmic}
\end{algorithm}
 
\begin{table}[!t]
\caption{Parameter values}
\vspace{-2mm}
\label{tab:exp_config}
\centering
\footnotesize
\setlength{\tabcolsep}{3pt}

\begin{tabularx}{\linewidth}{@{}lX@{}}
\toprule
\textbf{Item} & \textbf{Setting} \\
\midrule
Datasets & MNIST, CIFAR-10 \\
Clients / rounds & 100 / 500 ring rounds \\
Partition & IID; non-IID (Dirichlet $\beta=0.5$) \\
Optimizer & SGD, batch 80, $\eta_k=0.03/(1+0.03k)$ \\
Basil & $S=10$, 1 step; ACDS shares $5\%$ \\
FedSharp & $S=66$, $q=33$, $c=2.5$, $h=1$, $E=5$, $\mu=1$ \\
FedSharp threshold & $\texttt{Th}=c\sqrt{2\ln d}$ (11.94 MNIST, 12.08 CIFAR-10) \\
Byzantine & 33 clients; Gaussian $\mathcal{N}(\mathbf{0},\mathbf{I})$, hidden (from round 21) \\
Channel noise (std.) & $\sigma=0.01$ ($\sigma=0$ for reference) \\
\bottomrule
\end{tabularx}
\vspace{-2mm}
\end{table}

\begin{figure*}[htbp]
    \centering
    \includegraphics[width=\textwidth]{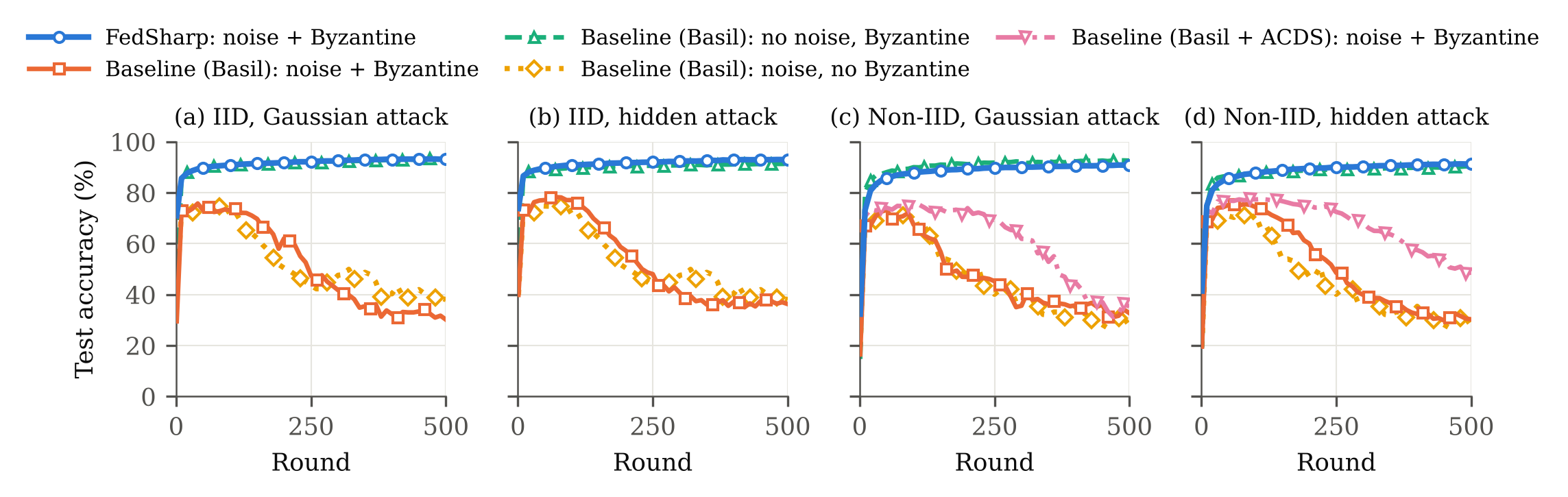}
    \caption{Accuracy comparison under Gaussian and hidden
    Byzantine attacks and noisy channel for MNIST IID and non-IID data settings.}
    \label{fig:mnist}
\end{figure*}
\begin{figure*}[htbp]
    \centering
    \includegraphics[width=\textwidth]{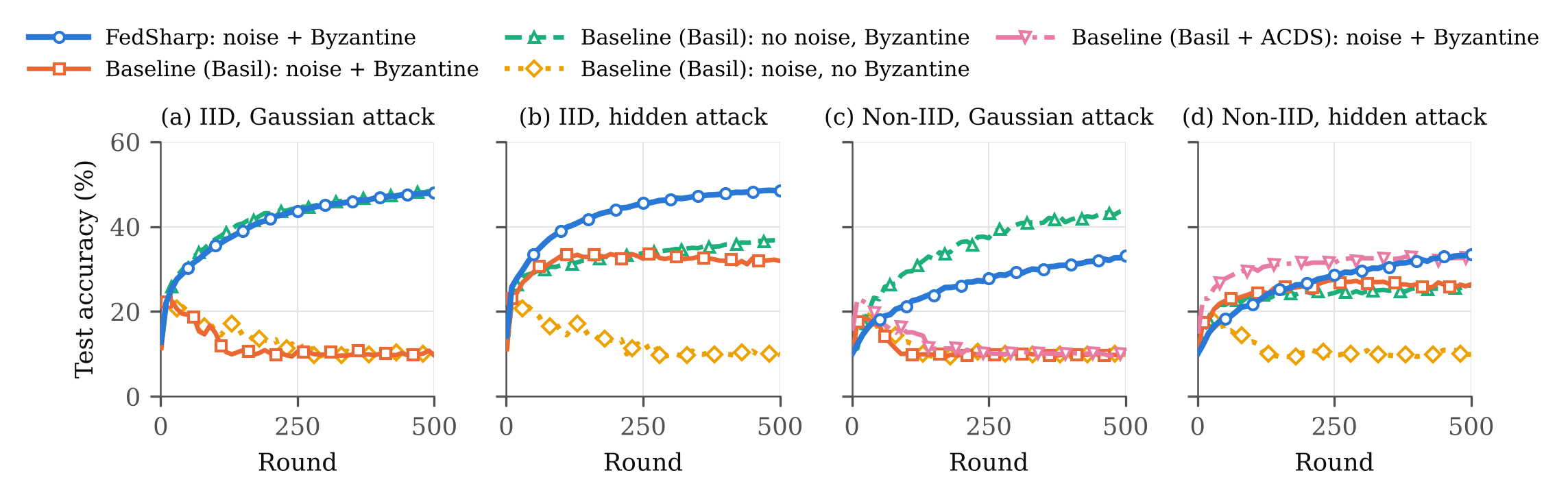}
    \caption{Accuracy comparison under Gaussian and hidden
    Byzantine attacks and noisy channel for CIFAR-10 IID and non-IID data settings.}
    \label{fig:cifar}
    \vspace{-2mm}
\end{figure*}

\section{Results and Discussion}
In this section, we evaluate the performance of FedSharp and compare it with the baseline (i.e., Basil and Basil+ACDS \cite{basil}) under the same impairments, channel noise ($\sigma=0.01$) and $b=33$ Byzantine nodes. Since ACDS is designed for heterogeneous data, we evaluate Basil+ACDS in the non-IID setting. Basil without channel noise and Basil without Byzantine nodes serve as references that isolate the effect of each impairment. The list of parameter values used in generating the results is summarized in Table~\ref{tab:exp_config}. 

Fig.~\ref{fig:mnist} presents the test accuracy over training rounds for Gaussian attacks (Figs.~\ref{fig:mnist}(a) and (c)) and hidden attacks (Figs.~\ref{fig:mnist}(b) and (d)), for IID data (Figs.~\ref{fig:mnist}(a) and (b)) and non-IID data (Figs.~\ref{fig:mnist}(c) and (d)).
 
Under the Gaussian attack, Basil initially improves but subsequently deteriorates under both IID and non-IID data. Since Basil rejects all malicious models, this degradation is primarily attributed to channel noise, as confirmed by the similar behavior without Byzantine nodes. Moreover, selecting a single model ($q=1$) leaves the noise floor $v^\ast\approx\sigma^2/(2q\eta_k h)$ unreduced as the learning rate decreases. In contrast, FedSharp maintains accuracy close to noise-free Basil. Although Basil+ACDS initially benefits from data sharing under non-IID conditions, it eventually deteriorates because ACDS does not mitigate channel noise. Consequently, FedSharp achieves $94.8\%$ (IID) and $93.5\%$ (non-IID), compared with $29.4\%$ and $35.0\%$ for Basil and $35.1\%$ for Basil+ACDS under non-IID data.

Under the hidden attack, Basil frequently selects malicious near-mean models because they contain less accumulated noise than honest received models. Although these models provide an unintended noise-filtering effect, Basil and Basil+ACDS still experience substantial performance degradation. In contrast, FedSharp's peak-based screening rejects most malicious models while preserving honest ones, limiting the attacker's contribution to approximately $2\%$. Consequently, FedSharp achieves $94.9\%$ (IID) and $93.8\%$ (non-IID), outperforming noise-free Basil and maintaining worst-honest-node accuracies of $94.7\%$ and $93.3\%$, respectively.
 
Fig.~\ref{fig:cifar} presents the CIFAR-10 results under Gaussian (Figs.~\ref{fig:cifar}(a) and (c)) and hidden attacks (Figs.~\ref{fig:cifar}(b) and (d)), for IID data (Figs.~\ref{fig:cifar}(a) and (b)) and non-IID data (Figs.~\ref{fig:cifar}(c) and (d)). Compared with MNIST, the larger network exhibits greater sensitivity to channel noise.

Under the Gaussian attack, Basil initially improves but eventually collapses to chance-level accuracy, even without Byzantine nodes, indicating that channel noise dominates performance degradation. In contrast, FedSharp maintains stable learning, achieving performance comparable to noise-free Basil under IID data. Under non-IID data, FedSharp converges more slowly but substantially outperforms Basil and Basil+ACDS, although it remains below noise-free Basil. This gap may reflect the combined effects of heterogeneous model averaging and proximal regularization, which can limit local adaptation.

Under the hidden attack, Basil performs better than under the Gaussian attack because the attackers' near-mean models partially suppress accumulated noise. Nevertheless, its performance stagnates, whereas FedSharp continues improving and outperforms Basil under both IID and non-IID data. Under IID data, FedSharp also exceeds noise-free Basil by mitigating inherited malicious perturbations through model screening. Under non-IID data, Basil+ACDS initially benefits from data sharing and the noise-filtering effect of near-mean attacker models, but FedSharp eventually surpasses it, achieving 33.0\% accuracy compared with 31.6\% for Basil+ACDS, without exchanging raw training data. This modest difference, however, is not sufficient to establish a statistically significant improvement from a single run.

\section{Conclusion}
This paper studied decentralized federated learning under three interacting challenges: non-IID data, Byzantine model corruption, and noisy communication channels. We introduced FedSharp, a robust decentralized learning framework that combines peak-based model screening, multi-model aggregation, hop-aware robust training, and proximal regularization to mitigate these impairments. Experimental results on MNIST and CIFAR-10 under Gaussian and hidden attacks demonstrated that FedSharp consistently outperforms Basil and Basil+ACDS in the evaluated settings, maintaining stable learning despite Byzantine attacks and channel noise. The results also highlight the importance of jointly addressing these challenges rather than treating them independently. However, performance degradation under strongly non-IID data remains an important direction for future research.
\vspace{-2mm}
\bibliographystyle{IEEEtran}
\bibliography{references_ieee}

\end{document}